\documentclass{article}

\usepackage[preprint, nonatbib]{nips_2018}

\usepackage{float}

\usepackage[square,numbers]{natbib}
\usepackage[utf8]{inputenc} 
\usepackage[T1]{fontenc}    
\usepackage{hyperref}       
\usepackage{url}            
\usepackage{booktabs}       
\usepackage{amsfonts}       
\usepackage{nicefrac}       
\usepackage{microtype}      
\usepackage{graphicx}
\usepackage{tabularx}
\usepackage{hyperref}
\usepackage{xcolor}

\title{A Proposal for Intelligent Agents\\with Episodic Memory}

\author{
  David Murphy \\
  HP Labs -- HP Inc.\\
  Palo Alto, CA, USA\\
  \texttt{david.murphy2@hp.com} \\
\And
  Thomas Paula \\
  HP Labs -- HP Inc.\\
  Porto Alegre, RS, Brazil\\
  \texttt{thomas.paula@hp.com} \\
\And
  Wagston Staehler \\
  HP Labs -- HP Inc.\\
  Porto Alegre, RS, Brazil\\
  \texttt{wagston@hp.com} \\
\And
  Juliano Vacaro \\
  HP Labs -- HP Inc.\\
  Porto Alegre, RS, Brazil\\
  \texttt{juliano.vacaro@hp.com} \\
\And
  Gabriel Paz \\
  HP Labs -- HP Inc.\\
  Porto Alegre, RS, Brazil\\
  \texttt{gabriel.lima.paz@hp.com} \\
\And
  Guilherme Marques \\
  HP Labs -- HP Inc.\\
  Porto Alegre, RS, Brazil\\
  \texttt{guilherme.marques@hp.com} \\
\And
  Bruna Oliveira \\
  HP Labs -- HP Inc.\\
  Porto Alegre, RS, Brazil\\
  \texttt{bruna.oliveira@hp.com} \\
}

\begin{document}
\maketitle

\begin{abstract}
In the future we can expect that artificial intelligent agents, once deployed, will be required to learn continually from their experience during their operational lifetime.
Such agents will also need to communicate with humans and other agents regarding the content of their experience, in the context of passing along their learnings, for the purpose of explaining their actions in specific circumstances or simply to relate more naturally to humans concerning experiences the agent acquires that are not necessarily related to their assigned tasks.
We argue that, to support these goals, an agent would benefit from an episodic memory; that is, a memory that encodes the agent's experience in such a way that the agent can relive the experience, communicate about it and use its past experience, inclusive of the agents own past actions, to learn more effective models and policies.
In this short paper, we propose one potential approach to provide an AI agent with such capabilities.
We draw upon the ever growing body of work examining the function and operation of the Medial Temporal Lobe (MTL) in mammals to guide us in adding an episodic memory capability to an AI agent composed of artificial neural networks (ANNs).
Based on that, we highlight important aspects to be considered in the memory organization and we propose an architecture combining ANNs and standard Computer Science techniques for supporting storage and retrieval of episodic memories.
Despite being initial work, we hope this short paper can spark discussions around the creation of intelligent agents with memory or, at least, provide a different point of view on the subject.
\end{abstract}

\section{Introduction}
The value of an episodic memory to aid learning in an artificial agent is gaining acceptance.
This is reflected in the increasing number of Deep Reinforcement Learning (DRL) papers in recent years that include an episodic memory in the agent architecture.
In a DRL based agent, this often takes the form of adding an external memory system, similar in architecture to the Differentiable Neural Computer (DNC)~\cite{graves2016hybrid}, to the agents policy network that decides which action or actions to take given the agents' observations of the current environment state~\cite{wayne2018unsupervised}.
The addition of these memories have both extended the horizon over which such agents can learn by mitigating the \textit{long horizon problem}~\cite{lillicrap2019backpropagation} and have decreased the number of episodes of experience an agent needs to learn an effective policy over agents that lack a memory.
 
The use of the term episodic memory in the DRL literature however, seems to differ from that of the cognitive sciences.
In DRL, the term often refers to the storage and retrieval of the hidden state of the neural network (often an LSTM~\cite{Hochreiter:1997:LSM:1246443.1246450} variant) implementing the policy of an agent~\cite{pmlr-v80-ritter18a}.
That is, an embedding of the recent past experience in the network's hidden layers that is preserved for later retrieval, allowing the agent to pick up from where it left off learning in the past.
In contrast to the usage in psychology and the cognitive sciences, the stored state is not decodable to recover the experience of an agent for a specific past episode.
On the other hand, Tulving~\cite{tulving1998episodic, Tulving2002} referred to episodic memory as a kind of mental time travel -- a way for an agent (in this case a human one) to recall an experience and move forward or backward in time within that experience to communicate about it or to reflect on it in service to another goal.
How could we construct a system that provides this class of memory capability to an AI agent?

\section{An Architecture supporting Storage and Retrieval of Episodic Memory}
A step towards answering this question might be found in examining mammalian episodic memory and extracting a set of properties to guide us in building an artificial analog.

The hippocampal formation (a part of the MTL) is thought by many to be involved in memory in general and episodic memory in particular~\cite{REAGH201869, Ranganath2012}.
An examination of the pathways in and out of the hippocampal formation show connections between it and both the Anterior-temporal (AT) network and the Posterior-medial (PM) network.
The former system is believed to extract and keep track of the objects and entities present in the world and their properties.
The PM network on the other hand keeps track of the relationships between the objects and entities.
The hippocampal formation records experience as reflected in activity of the AT and PM networks in such a way that when presented with a partial set of cues from either the relationships or the objects and entities, the complete set of objects, entities and relations from past experience are recalled, and this is accompanied by restoration of activity in the AT and PM networks.
In essence (perhaps oversimplifying), the activities of these networks is restored in such a way that the sensory experience of the memory is available.

We can extract some useful properties from this body of work that might inspire a similar capability in an artificial system.
First, the system is ``generative".
That is, from the encoding of the memory we can generate the representations that were present when the memory was stored.
Second, we can recall the complete memory (the objects, entities, and relationship present at the time of encoding) from a partial cue.
Third, while not explicitly called out above, we can say, with some certainty, that mammals maintain a belief about their current environment~\cite{ChanNivNorman2016}.
If we take these three characteristics together, we can begin to envision an artificial system that is both feasible and that would contain an episodic memory system.

\begin{figure}[!htb]
  \centering
  \includegraphics[width=0.6\linewidth]{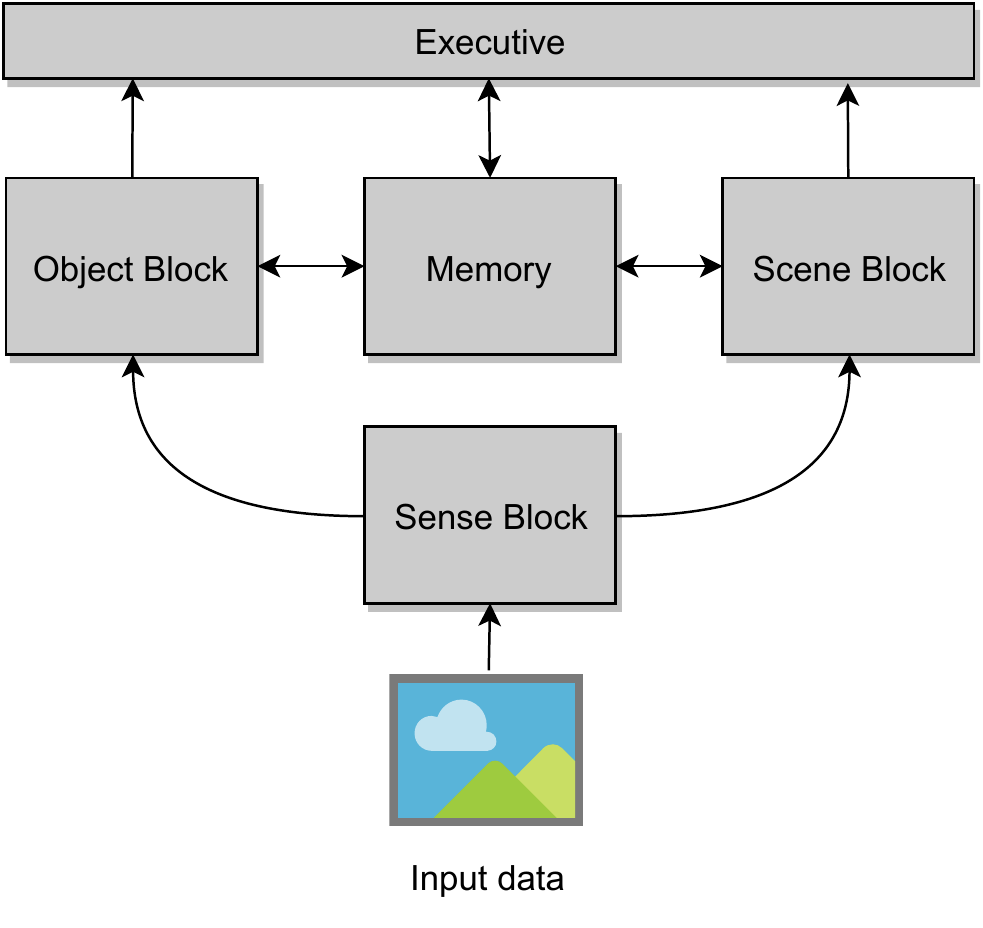}
  \caption{Simplified diagram of the proposed architecture.}
  \label{fig:simplified_architecture_diagram}
\end{figure}

In Figure~\ref{fig:simplified_architecture_diagram}, we see a simplified diagram upon which we explore how adhering to these principles we can construct such a system.
For simplification, we assume for the moment we proceed from visual input, but the ideas extend to any sensory modality.
Visual information comes into the \textit{sense block}, which extracts features vectors, $\phi$ from the image.
These feature vectors are consumed by an \textit{object block} and \textit{scene block}.
The object block is responsible for determining what entities are present in the environment from the feature vectors supplied by the sense block.
The object block is composed of two elements, an ANN that constructs a pose invariant feature vector $\phi_{inv}$ for each object represented in the feature vectors, and a \textit{working memory}.
The working memory maintains the current set of invariant feature vectors $\phi_{inv}$ representing the objects present in the agent's environment.
$\phi_{inv}$ are placed in working memory as they are produced, and decay from the memory if they are not refreshed by the future presentations of the $\phi_{inv}$.
The working memory provides temporal consistency that the sensory element does not.
The scene block construction is analogous to the object block.
It comprises an ANN that extracts relationships among objects, and stores those relationships in a working memory.

Together, the working memories of the object and scene blocks maintain the agent's belief about the world.
The set of objects and entities present and their relationships.
Once we have this belief, we can formulate the episodic memory as a system that tracks the agent's change in its belief.
That belief should change much more slowly than the feature vectors at the output of the sensory block for two reasons:
\begin{itemize}
  \item The invariant feature vectors change more slowly than the sensory feature vectors, as they are unaffected by pose; the object's or the agent's.
  \item The working memories provide some temporal consistency to the invariant feature vectors, even as the objects in view of the sense block that caused them appear and disappear from view temporarily from change in agent pose.
\end{itemize}
This slowly varying belief reduces the amount of storage required to capture experience over storing raw feature vectors.
 
It was stated earlier that the biological system is generative.
If we place this constraint on the ANNs that make up the sense, object and scene blocks, then we can view recall as reinstating the working memory from the changes in belief stored in the episodic memory and relive, so to speak, the sensory experience by regenerating the feature vectors of the sense block from those of the working memory.
An agent then has the option of moving through these stored beliefs, forward or backward in time providing a facility akin to Tulving's.
Note that the degree of reconstruction required would depend on the intended use.
A DRL agent might just require the highest level abstract features as input to policy updates whereas a response to presenting the query \textit{``Show me where you were\dots"} to an agent might be a pixel level reconstruction of the experience that would be suitable for a person to consume.

The executive block in Figure~\ref{fig:simplified_architecture_diagram} represents the policy network of an artificial agent.
The policy network receives input from the object and scene blocks as its perceptual inputs but also queries the episodic memory in service to the policy.

\section{Memory Organization}
We have stated that the episodic memory system tracks the changes in the agent's belief but we have not articulated how such a memory is organized and how information is stored and retrieved.
For the memory system to be effective, we must satisfy three properties.
First, we must be able to store an accurate memory from a single exposure to the environment; second, to retrieve the memory from a partial cue; and third, we must be able to store and recall a large amount of experience.

There is considerable recent work on augmenting deep neural networks with external memory~\cite{SukhbaatarSWF15, wayne2018unsupervised, weston2014memory, Kumar2015, pritzel2017neural, sodhani2020toward, badia2020agent57}.
In common formulations, access to the memory requires a scan over the entire memory while applying some function to the contents that ultimately computes a weighting that identifies which memories are of most value~\cite{GravesWD14}.
Sweeping over the entire memory may not be unreasonable biologically, perhaps, as all the computing elements and memory can operate (theoretically) in parallel, but it is unreasonable for our current computer systems as either the energy, computation or time required would become untenable.
Therefore, organizing the storage of memory and the representations stored there in such a way that past memory is retrievable with bounded computation and that similar memories are nearby as defined by the search mechanism is required.
 
We can achieve these three properties if we impose an additional constraint on the representations produced and placed in the working memory.
That constraint is that the representation is composable, or disentangled~\cite{bengio2013representation}.
With this constraint, value of a dimension of the disentangled feature vector of a set of objects or relationships for which a property is common will be similar.
In other words similar objects and relationships would lie near each other in the feature vector space.

\begin{figure}[!htb]
  \centering
  \includegraphics[width=1.0\linewidth]{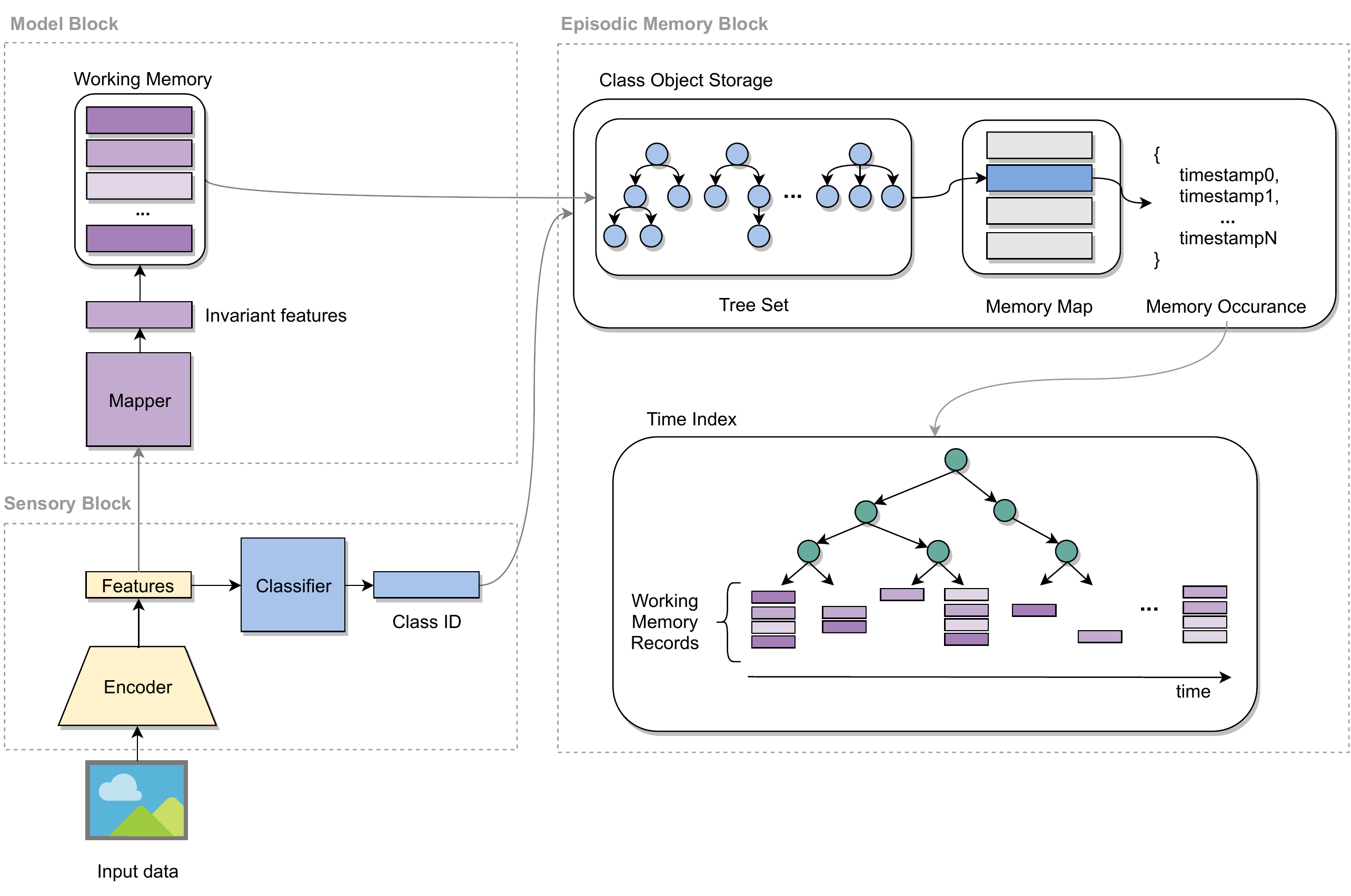}
  \caption{High-level diagram of the proposed system, showing the subset for objects. The Sensory block produces a disentagled feature vector which passed on to the Model block which produces a feature vector invariant to the agents viewpoint $\phi_{inv}$. The $\phi_{inv}$ is maintained in working memory providing an agents belief about the state of the world. Changes to the working memory are stored in the Episodic Memory block}
  \label{fig:architecture_diagram}
\end{figure}

With this additional property, we can cast the memory into an assembly of classic computer science approaches to storage (see Figure~\ref{fig:architecture_diagram}).
We define a \textit{memory occurance} data structure to store each instance of an object or relationship class.
This data structure maintains a copy of the invariant feature vector $\phi_{inv}$, and a list of timestamps when the object or relation was entered or removed from working memory.
We view each $\phi_{inv}$ as a point in the feature space corresponding to a particular object or relation class.
Within a class, in order to quickly determine if a particular instance of a class has been seen, an index into the feature space is constructed by assigning an ordered tree structure to each dimension of the feature vector where each leaf node corresponds to a value of a dimension.
The collection of trees corresponding to the dimensions of the feature vector we refer to as a \textit{tree set}.
When storing a change in the agents belief, for each new or deleted feature vector, we search the tree set for a match.
If we find one we know this class instance has been seen before and we construct a hash from the IDs of the leaf nodes (each node in the tree having a unique ID) and use it to index the data structure that maintains the history of this specific object or relationship instance.
If we do not find one then we have not seen this instance before.
Consequently, we create a new entry in the tree set and in the hashtable.

While the above mechanism solves storage and retrieval of an individual object or relationship, it is does not solve recovery of the complete memory.
To achieve this, we maintain a running list of changes to the working memory.
This running list of working memory changes is ordered in time and we can rely on this property to build efficient indexing into the memory.
Hence, once we recovery a partial memory by a search in the tree set, we can recover the time index for that memory and retrieve and reconstruct the entire working memory corresponding to the agent's belief at that time.

Earlier we stated that we must be able to retrieve a memory from a partial cue.
Partial cues take two forms.
An object or relationship that is part of a complete memory or an under-specified description of an object or relationship (e.g. a cat vs a cat with black fur and a white underbelly with yellow eyes).
The former is addressed in the description above.
To address the latter we can consider one form of a partial cue as neglecting some of the dimensions of the feature vector when searching the tree set.
For example, we might neglect color, and in so doing retrieve all objects of all possible colors that match the rest of the specification.
 
Furthermore, since the tree set structure is ordered, we can bound searches in the tree set for elements that have similar properties.
That is, since we know that increasing (or decreasing) a particular value in a particular dimension corresponds to decreasing similarity we can bound similarity search by limiting how much change in value of a particular dimension is appropriate for our search.

\section{Discussion}
We have described an approach to adding episodic memory to artificial agents that we arrived at by drawing inspiration from the large body of work on the Medial Temporal Lobe (MTL) and how the MTL interacts with other brain areas.
The properties enabled by the MTL guide us in formulating an analog for an system composed of ANNs.
We show that, by carefully restricting the properties of the ANNs involved and adding a working memory representing an agent's belief about the world -- temporal consistency of representation -- that we can cast episodic memory as augmenting an ANN system with well know computer science approaches to memory.

The systems level approach taken here means that no individual component stands in isolation.
Rather, the properties of each component are shaped by the system architecture and operation. In our case this lead us to make important assumptions regarding the features the individual components provide: disentanglement and invariance.
Bengio et al.\cite{bengio2013representation} state, as is often the case in the context of deep learning methods, that the feature set produced in training may be destined for use in multiple tasks that may each rely on distinct subsets of relevant features. 
The authors conclude that the most robust approach to feature learning is to disentangle as many factors as possible, discarding as little information about the data as is practical. 
Disentanglement has been the subject of study of recent work and various approaches to learning disentangled features have been developed~\cite{greff2019multi, watters2019spatial}. The property is useful for memory as it allows for matching on subsets of the features but also because it allows for composable representations.
Models that extract disentangled features remains an open research topic, as are methods for evaluating the quality of the representations. In Locatello et al.~\cite{locatello2018challenging}, the authors argue that unsupervised learning of disentangled representations might only be possible with inductive biases on both the models and the data.
While methods for producing disentangled features is an open research problem, we nonetheless believe this property of features is useful as it allows for search by partially specified queries.

Feature invariance is another important assumption we make.
The working memory proposed relies on invariant features produced by a given model such that  objects or relationships can be matched to their feature vectors in working memory across gaps in their appearance.
The proposed system should be able to extract invariant features from the disentangled representations which are saved  ultimately to episodic memory.
The generative nature of the system when recalling memories means such a system should be able to map invariant features back to a representation that allows the decoder portion of the VAE in the sensory block to recreate the input.
The authors acknowledge that the imposition of these system level constrains of obtaining both disentanglement and invariance is challenging and not to be underestimated, but believe the constraints greatly simplify the memory architecture.

We built a preliminary system and have working components for a subset of the system: sensory, model, and episodic memory blocks (Figure~\ref{fig:architecture_diagram}).
We are able to store and retrieve memories, and early results seem promising, but we have work to do on improving the representations the ANNs produce.
In our initial experiments, we used a generative model based on variational autoencoders (VAEs)~\cite{kingma2016improved} and explored the use of techniques to improve disentanglement such as the spatial broadcast decoder~\cite{watters2019spatial}.
For the model block, we conducted experiments using contrastive learning~\cite{chopra2005learning} to train an ANN to produce the invariant representations.
The loss function encourages similar entities to have similar representations and different entities to have different representations~\cite{schroff2015facenet}.
Of course, considering all of the elements together is challenging, but our initial results are encouraging.

As a final thought, previous work in human subjects show the relationship of episodic memory to imagination~\cite{Hassabis1726}.
Subjects with damage to the hippocampus, and hence impaired episodic memory capabilities also produce less rich imagined scenes.
The type of system described here may also admit a form of imagination.
That is, as we have organized memory so that similar objects and relations appear close to one another in feature space, and that the representations are composable.
We can construct an imagined memory by sampling the memory space with some amount of induced noise. 
This imagined memory will be composed of elements similar to the agents previous experience.

We hope the ideas expressed here can spark discussion on memory and its place in artificial systems and bring us closer to agents that learn over their lifetime.
We plan to conduct additional experiments using the proposed architecture and share results with the research community.

\bibliography{references}

\end{document}